\documentclass[11pt,a4paper]{article}
\usepackage[hyperref]{acl2017}
\usepackage{times}
\usepackage{latexsym}
\usepackage{url}
\usepackage{adjustbox}
\usepackage{tikz}
\usepackage{tikz-dependency}
\usepackage{qtree}
\usepackage{subcaption}
\usepackage{enumitem}

\usepackage{amsmath}
\usepackage{verbatim}
\usepackage{tikz-qtree}
\usepackage{tikz}

\usepackage{filecontents}
\usepackage{pgfplots, pgfplotstable}
\usepgfplotslibrary{statistics}

\usetikzlibrary{shapes.geometric}
\usetikzlibrary{shapes.arrows}
\usetikzlibrary{patterns}
\usetikzlibrary{chains}
\usetikzlibrary{calc}

\aclfinalcopy 

\newcommand{\ignore}[1]{}

\qtreecenterfalse

\title{Arc-Standard Spinal Parsing with Stack-LSTMs}

\author{Miguel Ballesteros \\ IBM T.J Watson Research Center \\ Yorktown Heights, NY 10598. U.S. \\ miguel.ballesteros@ibm.com
         \And  
        Xavier Carreras \\ Naver Labs Europe \\ Meylan, France \\ xavier.carreras@naverlabs.com}
\date{}

\begin{document}
\maketitle

\begin{abstract}
  We present a neural transition-based parser for spinal trees, a
  dependency representation of constituent trees. The parser uses
  Stack-LSTMs that compose constituent nodes with dependency-based
  derivations. In experiments, we show that this model adapts to
  different styles of dependency relations, but this choice has little
  effect for predicting constituent structure, suggesting that LSTMs
  induce useful states by themselves.
  \end{abstract}

\section{Introduction}

There is a clear trend in neural transition systems 
for parsing sentences into dependency trees
\cite{titov-henderson:2007:IWPT2007,chen2014fast,Dyer:ACL15,andor2016} and constituent
trees
\cite{henderson:2004:ACL,vinyals,watanabe-sumita:2015:ACL-IJCNLP,dyer2016,Cross:EMNLP16}.
These transition systems use a relatively simple set of operations to
parse  in linear time, and rely on the ability of neural
networks to infer and propagate hidden structure through the
derivation. This contrasts with state-of-the-art factored linear
models, which explicitly use of higher-order information to capture
non-local phenomena in a derivation. 

In this paper, we present a transition system for parsing sentences
into spinal trees, a type of syntactic tree that explicitly represents
together dependency and constituency structure.
This representation is inherent in head-driven models
\cite{collins:1997:ACL} and was used by \citet{Carreras:CoNLL08}
with a higher-order factored model.
We extend the Stack-LSTMs by \citet{Dyer:ACL15} from dependency to
spinal parsing, by augmenting the composition operations to include
constituent information in the form of spines. To parse sentences, we
use the extension by \newcite{CrossHuang16a} of the arc-standard system
for dependency parsing \cite{Nivre:2004}. This parsing system
generalizes shift-reduce methods
\cite{Henderson2003,sagae-lavie:2005:IWPT,zhu-EtAl:2013:ACL20131,watanabe-sumita:2015:ACL-IJCNLP}
to be sensitive to constituent heads, as opposed to, for example,
parse a constituent from left to right.

In experiments on the Penn Treebank, we look at how
sensitive our method is to different styles of dependency
relations, and show that spinal models based on leftmost or rightmost
heads are as good or better than models using linguistic dependency
relations such as Stanford Dependencies \cite{deMarneffe} or those by \citet{yamada2003statistical}.
This suggests that Stack-LSTMs figure out effective ways of modeling
non-local phenomena within constituents. We also show that turning a
dependency Stack-LSTM into spinal results in some
improvements.


\section{Spinal Trees}

\begin{figure}[t]
  \def\www{0.9}
  \begin{subfigure}{\linewidth}
  \centering
  \adjustbox{width=\www\linewidth}{    
  \Tree [.S And [.NP [.NP their suspicions ] [.PP of [.NP each other ] ] ]  [.VP run [.ADVP deep ] ]   $\cdot$ ]      
   }
  \caption{A constituency tree from the Penn Treebank.}
  \label{fig:ctree}
  \end{subfigure}  
  \vskip 1em
  
  \begin{subfigure}{\linewidth}
  \centering
  \adjustbox{width=\www\linewidth}{
    \input{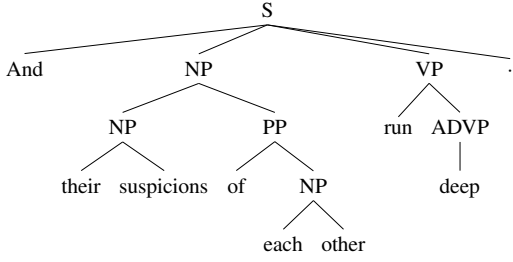}
  }
  \caption{The spinal tree of (\ref{fig:ctree}) using Stanford Dependency heads.}
  \label{fig:stree-sd}
  \end{subfigure}  
  \vskip 1em

  \begin{subfigure}{\linewidth}
  \centering
  \adjustbox{width=\www\linewidth}{
    \input{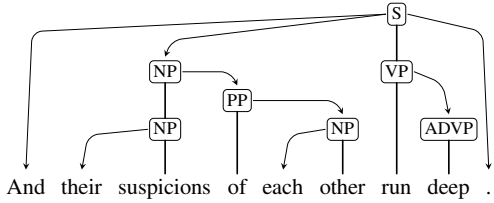}
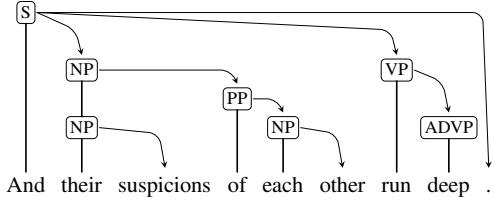
  }
  \caption{The spinal tree of (\ref{fig:ctree}) using leftmost heads.}
 \label{fig:stree-left}
  \end{subfigure}  

  \caption{A constituency tree and two spinal trees.}
  \label{fig:trees}
\end{figure}

%
In a spinal tree each token is associated with a spine. The spine of a
token is a (possibly empty) vertical sequence of non-terminal nodes
for which the token is the head word. A spinal dependency is a binary
directed relation from a node of the head spine to a dependent
spine. In this paper we consider projective spinal trees.
Figure \ref{fig:trees} shows a constituency tree from the Penn Treebank
together with two spinal trees that use alternative head identities:
the spinal tree in \ref{fig:stree-sd} uses Stanford Dependencies
\cite{deMarneffe}, while the spinal tree in \ref{fig:stree-left} takes
the leftmost word of any constituent as the head.
It is direct to map a constituency tree with head annotations
to a spinal tree, and to map a spinal tree to a constituency or a
dependency tree.

\section{Arc-Standard Spinal Parsing}
\label{arcstd}

\newcommand{\sone}[2]{#1\!\mathop{:}\!#2}
\newcommand{\stwo}[3]{#1\!\mathop{:}\!#2\!\mathop{:}\!#3}
\newcommand{\splus}{\raisebox{0.5pt}{\scalebox{0.8}{+}}}
\newcommand{\dplus}{\raisebox{0.5pt}{\scalebox{0.6}{$\cup$}}}

%

We use the transition system by \citet{CrossHuang16a}, which
extends the arc-standard system by \citet{Nivre:2004} for constituency
parsing in a head-driven way, i.e. spinal parsing. We describe it here for completeness. 
The parsing state is a tuple
\mbox{$\langle \beta, \sigma, \delta \rangle$}, where $\beta$ is a
buffer of input tokens to be processed; $\sigma$ is a stack of tokens
with partial spines; and $\delta$ is a set of \emph{spinal}
dependencies.
The operations are the following:
\begin{itemize}
\item {\tt shift} : $\langle \sone{i}{\beta},\, \sigma,\,
  \delta\rangle \to \langle \beta,\, \sone{\sigma}{i},\, \delta
  \rangle$
  
  Shifts the first token of the buffer $i$ onto the stack,
  $i$ becomes a base spine consisting of a single token.

\item {\tt node($n$)} :
  $\langle \beta,\, \sone{\sigma}{s},\, \delta\rangle \to
  \langle \beta,\, \sone{\sigma}{s\splus n},\, \delta \rangle$ 
  
  Adds a non-terminal node $n$ onto the top element of the stack $s$,
  which becomes $s\splus n$. At this point, the node $n$ can receive
  left and right children (by the operations below) until the node is
  closed (by adding a node above, or by reducing the spine with an arc
  operation with this spine as dependent). By this single operation,
  the arc-standard system is extended to spinal parsing.
  
\item {\tt left-arc} : \\
  $\langle \beta,\, \stwo{\sigma}{t}{s\splus n},\, \delta\rangle \to
  \langle \beta,\, \sone{\sigma}{s \splus n},\, \delta \dplus (n,t) \rangle$ 

  The stack must have two elements, the top one is a spine $s \splus
  n$, where $n$ is the top node of that spine, and the second element
  $t$ can be a token or a spine.  The operation adds a spinal from the
  node $n$ to $t$, and $t$ is reduced from the stack. The dependent
  $t$ becomes the leftmost child of the constituent $n$.
  
\item {\tt right-arc} : \\
  $\langle \beta,\, \stwo{\sigma}{s \splus n}{t},\, \delta\rangle \to
  \langle \beta,\, \sone{\sigma}{s \splus n},\, \delta \dplus (n,t) \rangle$
  
  This operation is symmetric to {\tt left-arc}, it adds a spinal
  dependency from the top node $n$ of the second spine in the stack to the top
  element $t$, which is reduced from the stack and becomes the
  rightmost child of $n$.
 \end{itemize}

\begin{figure}[t]
  \centering  
  \adjustbox{width=\linewidth}{
\begin{tabular}{llll}
  {\bf Transition}  & {\bf Buffer $\beta$} & {\bf Stack $\sigma$} &  {\bf New Arc in $\delta$}\\
  \hline
  & [And, their, \ldots] & [] & \\
  {\tt shift} & [their, suspicions, \ldots] & [And] & \\
  {\tt shift} & [suspicions, of, \ldots] & [And, their] & \\
  {\tt shift} & [of, each, \ldots] & [\ldots, their, susp.] & \\
  {\tt node(NP)} & [of, each, \ldots] & [\ldots, their, susp.\splus NP$_3^1$] & \\
  {\tt left-arc} & [of, each, \ldots] & [And, susp.\splus NP$_3^1$] & (NP$_3^1$,their)\\
  {\tt node(NP)} & [of, each, \ldots] & [And, susp.\splus NP$_3^1$\splus NP$_3^2$] & \\
  {\tt shift} & [each, other, \ldots] & [\ldots, susp.\splus NP$_3^1$\splus NP$_3^2$, of] & \\
  {\tt node(PP)} & [each, other, \ldots] & [\ldots, susp.\splus NP$_3^1$\splus NP$_3^2$, of\splus PP$_4^1$] & \\
  {\tt shift} & [other, run, \ldots] & [\ldots, of\splus PP$_4^1$, each] & \\
  {\tt shift} & [run, deep, \ldots] & [\ldots, each, other] & \\
  {\tt node(NP)} & [run, deep, \ldots] & [\ldots, each, other\splus NP$_6^1$] & \\
  {\tt left-arc} & [run, deep, \ldots] & [\ldots, of\splus PP$_4^1$, other\splus NP$_6^1$] & (NP$_6^1$, each)\\
   {\tt right-arc} & [run, deep, \ldots] & [\ldots, susp.\splus NP$_3^1$\splus NP$_3^2$, of\splus PP$_4^1$] & (PP$_4^1$, NP$_6^1$)\\
   {\tt right-arc} & [run, deep, \ldots] & [And, susp.\splus NP$_3^1$\splus NP$_3^2$] & (NP$_3^2$, PP$_4^1$)\\
   \ldots \\
\end{tabular}
}

\caption{Initial steps of the arc-standard derivation for the spinal
  tree in Figure \ref{fig:stree-sd}, until the tree headed at
  ``suspicions'' is fully built.  Spinal
  nodes are noted $n_i^j$, where $n$ is the non-terminal, $i$ is the
  position of the head token, and $j$ is the node level in the spine.}
\label{fig:derivation}
\end{figure}

At a high level, the system builds a spinal tree headed at token $i$
by:
\begin{enumerate}[nolistsep]
\item Shifting the $i$-th token to the top of the stack. By induction, 
the left children of $i$ are in the stack and are complete.
\item \label{enum:node} Adding a constituency node $n$ to $i$'s spine.
\item \label{enum:left-arc} Adding left children to $n$ in head-outwards order with {\tt
  left-arc}, which are removed from the stack.
\item \label{enum:right-arc} Adding right children to $n$ in head-outwards order with {\tt right-arc}, which are built recursively.
\item Repeating steps \ref{enum:node}-\ref{enum:right-arc} for as many
  nodes in the spine of $i$.
\end{enumerate}
Figure \ref{fig:derivation} shows an example of a derivation. The
process is initialized with all sentence tokens in the buffer, an
empty stack, and an empty set of dependencies. Termination is always
attainable and occurs when the buffer is empty and there is a single
element in the stack, namely the spine of the full sentence head.
This transition system is correct and sound with respect to the class
of projective spinal trees, in the same way as the arc-standard system
is for projective dependency trees \cite{Nivre:2008}. A derivation has
$2n + m$ steps, where $n$ is the sentence length and $m$ is the number
of constituents in the derivation.

We note that the system naturally handles constituents of arbitrary
arity. In particular, unary productions add one node in the spine
without any children. In practice we put a hard bound on the number of
consecutive unary productions in a spine\footnote{Set to 10 in our
  experiments}, to ensure that in the early training steps the model
does not generate unreasonably long spines.
We also note there is a certain degree of non-determinism: left and right arcs (steps
\ref{enum:left-arc} and \ref{enum:right-arc}) can be mixed as long as
the children of a node are added in head-outwards order. At training
time, our \emph{oracle derivations} impose the order above (first left
arcs, then right arcs), but the parsing system runs freely.
Finally, the system can be easily extended with dependency labels, but
we do not use them.


%


%

\section{Spinal Stack-LSTMs}


\citet{Dyer:ACL15} presented an arc-standard parser that uses
Stack-LSTMs, an extension of LSTMs \cite{hochreiter:1997} for transition-based systems
that maintains an embedding for each element in the stack.\footnote{We
  refer interested readers to
  \cite{Dyer:ACL15,Ballesterosetal2017}.}. Our model is based on the
same architecture, with the addition of the {\tt node($n$)} action.
The state of our algorithm presented in Section \ref{arcstd} is
represented by the contents of the \textsc{stack}, the \textsc{buffer} and
a list with the history of actions with Stack-LSTMs.  This state
representation is then used to predict the next action to take.

%


\paragraph{Composition:} when the parser predicts a {\tt left-arc()} or {\tt right-arc()}, we compose the vector
representation of the head and dependent elements; this is 
equivalent to what it is presented by  \citet{Dyer:ACL15}. The representation is obtained recursively as follows:
\begin{align*}
  \mathbf{c} = \tanh\left(\mathbf{U}[\mathbf{h}; \mathbf{d}] + \mathbf{e} \right). 
\end{align*}

\noindent where $\mathbf{U}$ is a learned parameter matrix, \textbf{h}
represents the head spine and \textbf{d} represents the dependent
spine (or token, if the dependent is just a single token) ; \textbf{e}
is a bias term.


Similarly, when the parser predicts a {\tt node($n$)} action, we
compose the embedding of the non-terminal symbol that is added
($\mathbf{n}$) with the representation of the element at the top of
the stack ($\mathbf{s}$), that might represent a spine or a single
terminal symbol. The representation is obtained recursively as
follows:
\begin{equation}
\mathbf{c} = \tanh\left(\mathbf{W}[\mathbf{s}; \mathbf{n}] + \mathbf{b} \right). \label{eq:n-comp}
\end{equation}
\noindent where $\mathbf{W}$ is a learned parameter matrix, \textbf{s} represents the token in the stack (and its partial spine, if non-terminals have been added to it) and \textbf{n} represents the non-terminal symbol that we are adding to \textbf{s}; \textbf{b} is a bias term.


As shown by \citet{kuncoroeacl} composition is an essential component in this kind of  parsing models.

\section{Related Work}

%
\citet{collins:1997:ACL} first proposed head-driven derivations for
constituent parsing, which is the key idea for spinal parsing, and
later \citet{Carreras:CoNLL08} came up with a higher-order graph-based
parser for this representation. 
Transition systems for spinal parsing are not new.
\citet{Ballesteros:CoNLL15} presented an arc-eager system that labels
dependencies with constituent nodes, and builds the spinal tree in
post-processing. \citet{Hayashi2016J} and \citet{Hayashi:ACL16} presented a
bottom-up arc-standard system that assigns a full spine with the {\tt
  shift} operation, while ours builds spines incrementally and does
not depend on a fixed set of full spines. Our method is different from
shift-reduce constituent parsers
\cite{Henderson2003,sagae-lavie:2005:IWPT,zhu-EtAl:2013:ACL20131,watanabe-sumita:2015:ACL-IJCNLP}
in that it is head-driven. \citet{CrossHuang16a} extended the
arc-standard system to constituency parsing, which in fact corresponds
to spinal parsing. The main difference from that work relies on the
neural model: they use sequential BiLSTMs, while we use Stack-LSTMs and composition functions.
Finally, dependency parsers have been extended to constituency parsing by encoding the additional structure in the dependency labels, in different ways \citep{Hall2007,Hall2008,FernandezGonzalez:ACL15}. 


\section{Experiments}

\begin{table}
 \centering
\adjustbox{max width = \linewidth}{
\begin{tabular}{|lcccc|}
\hline
      & LR    & LP    & F1   & $\!\!\!\!\text{UAS}\ _\text{(SD)}\!\!$ \\
\hline
Leftmost heads & 91.18 & 90.93 & 91.05 & -\\
Leftmost h., no n-comp & 90.20 & 90.76 & 90.48 & -\\
%
Rightmost heads  & 91.03 & 91.20 & 91.11 & - \\
%
Rightmost h., no n-comp  & 90.64 & 91.24 & 90.04 & - \\
SD heads & 90.75 & 91.11 & 90.93  & 93.49 \\
SD heads, no n-comp & 90.38 & 90.58 & 90.48  & 93.16 \\
SD heads, dummy spines & - & - & - &  93.30 \\
YM heads & 90.82 & 90.84 & 90.83 & - \\
\hline
\end{tabular}
}
\caption{Development results for spinal models, in terms of labeled
  precision (LP), recall (LR) and F1 for constituents, and unlabeled
  attachment score (UAS) against Stanford dependencies. Spinal models
  are trained using different head annotations (see text). Models
  labeled with \emph{``no n-comp''} do not use node compositions. The
  model labeled with \emph{``dummy spines''} corresponds to a standard dependency model.}
\label{tab:ctrees-dev}
\end{table}

We experiment with stack-LSTM spinal models trained with different
types of head rules. Our goal is to check how the head identities,
which define the derivation sequence, interact with the ability of
Stack-LSTMs to propagate latent information beyond the local scope of
each action. We use the Penn Treebank \cite{penntreebank} with
standard splits.\footnote{We use the the same POS tags as
  \citet{Dyer:ACL15}.}


We start training four spinal models, varying the head rules that define the spinal derivations:\footnote{It is simple to obtain a spinal tree given a constituency tree and a corresponding dependency tree. We assume that the dependency tree is projective and nested within the constituency tree, which holds for the head rules we use.}
\begin{itemize}[noitemsep]
\item Leftmost heads as in Figure \ref{fig:stree-left}.
\item Rightmost heads.
\item Stanford Dependencies (SD) \cite{deMarneffe}, as in Figure \ref{fig:stree-sd}.
\item Yamada and Matsumoto heads (YM) \cite{yamada2003statistical}.
\end{itemize}

Table \ref{tab:ctrees-dev} presents constituency and dependency metrics on the development set. 
The model using rightmost heads works the best at 91.11 F1, followed
by the one using leftmost heads. It is worth to note that the two models using structural
head identities (right or left) work better than those using
linguistic ones. This suggests that the Stack-LSTM model already finds
useful head-child relations in a constituent by parsing from the left
(or right) even if there are non-local interactions. In this case,
head rules are not useful.

The same Table \ref{tab:ctrees-dev} shows two ablation studies. First,
we turn off the composition of constituent nodes into the latent
derivations (Eq \ref{eq:n-comp}). The ablated models, tagged with
\emph{``no n-comp''}, perform from 0.5 to 1 points F1 worse, showing
the benefit of adding constituent structure.
Then, we check if constituent structure is any useful for dependency
parsing metrics. To this end, we emulate a dependency parser using a
spinal model by taking standard Stanford dependency trees and adding a
dummy constituent for every head with all its children. This model, tagged ``SD
heads, dummy spines'', is slightly outperformed by the ``SD heads'' model
using true spines, even though the margin is small.

Tables \ref{tab:ctrees-test} and \ref{tab:dtrees} present results
on the test, for constituent and dependency parsing respectively. As shown in Table \ref{tab:ctrees-test} our model is competitive compared to the best parsers; the generative parsers by \citet{Choe:EMNLP16}, \citet{dyer2016} and \citet{kuncoroeacl} are better than the rest, but compared to the rest our parser is at the same level or better. The most similar system is by \citet{Ballesteros:CoNLL15} and our parser significantly improves the performance.
Considering dependency parsing, our model is worse than the ones that train with exploration as \citet{kiperwasser} and \citet{ballesteros2016}, but it slightly improves the parser by \citet{Dyer:ACL15} with static training. The systems that calculate dependencies by transforming phrase-structures with conversion rules and that use generative training are ahead compared to the rest.


\begin{table}
\centering
\adjustbox{max width = \linewidth}{
\begin{tabular}{|l|ccc|}
\hline
& LR & LP & F1 \\
\hline
Spinal (leftmost) & 90.30 & 90.54 & 90.42 \\
Spinal (rightmost) & 90.23 & 90.77 & 90.50 \\
\hline 
\citet{Ballesteros:CoNLL15} & 88.7 & 89.2& 89.0\\
\citet{vinyals} {\small (PTB-Only)} & & & 88.3 \\ 
\citet{CrossHuang16a}       &  &  & 89.9 \\
\newcite{chocharniak} {\small(PTB-Only)} & & & 91.2 \\
\newcite{chocharniak} {\small(Semi-sup)} & & & 93.8 \\
\citet{dyer2016} {\small (Discr.)} & & & 91.2 \\
\citet{dyer2016} {\small (Gen.)} & & & 93.3 \\
\newcite{kuncoroeacl} {\small(Gen.)} & & & 93.5 \\
\citet{LiuZhang2017} & 91.3 & 92.1 & 91.7 \\
\hline
\end{tabular}
}
\caption{Constituency results on the PTB test set.}
\label{tab:ctrees-test}
\end{table}

\begin{table}
\centering
\adjustbox{max width=\linewidth}{
\begin{tabular}{|l|c|}
\hline
   & UAS test \\
\hline
Spinal, PTB spines + SD (TB-greedy) &  93.15 \\
Spinal, dummy spines + SD (TB-greedy) &  93.10 \\
\hline
\citet{Dyer:ACL15} (TB-greedy) & 93.1 \\
\citet{CrossHuang16a}          & 93.4     \\
\citet{ballesteros2016} (TB-dynamic)  & 93.6 \\
\citet{kiperwasser} (TB-dynamic) &  93.9 \\ 
\newcite{andor2016} (TB-Beam) &  94.6 \\ 
\newcite{kuncoroEMNLP2016} (Graph-Ensemble) & 94.5 \\ 
\newcite{chocharniak}* (Semi-sup) &  95.9 \\
\newcite{kuncoroeacl}* (Generative)  & 95.8 \\
\hline
\end{tabular}
}
\caption{Stanford Dependency results (UAS) on PTB test set. Parsers marked with * calculate dependencies by transforming phrase-structures with conversion rules.}
\label{tab:dtrees}
\end{table}

\section{Conclusions}

We have presented a neural model based on Stack-LSTMs for spinal
parsing, using a simple extension of arc-standard transition parsing
that adds constituent nodes to the dependency derivation.  Our
experiments suggest that Stack-LSTMs can figure out useful internal
structure within constituents, and that the parser might work better
\emph{without} providing linguistically-derived head words.
Overall, our spinal neural method is simple, efficient, and very
accurate, and might prove useful to model constituent trees with
dependency relations.


\bibliography{refs}
\bibliographystyle{acl_natbib}

\end{document}